\documentclass[runningheads]{llncs}
\usepackage{graphicx}
\usepackage{amsmath,amssymb} 
\usepackage{color}
\usepackage{tabularx, booktabs}
\graphicspath{ {img/} }

\newcolumntype{Y}{>{\centering\arraybackslash}X}

\newcommand{\argmin}{\operatornamewithlimits{argmin}}

\begin{document}

\newcommand{\point}{
    \raise0.7ex\hbox{.}
    }


\pagestyle{headings}

\mainmatter

\title{Dense Motion Estimation for Smoke} 

\titlerunning{Dense Motion Estimation for Smoke} 

\authorrunning{Da Chen, Wenbin Li, Peter Hall} 

\author{Da Chen$^{1}$ \ \ \ Wenbin  Li$^{2}$ \ \ \ Peter Hall$^{1}$} 
\institute{$^{1}$Department of Computer Science, University of Bath \\ $^{2}$Department of Computer Science, University College London
} 

\maketitle

\begin{abstract}
Motion estimation for highly dynamic phenomena such as smoke is an open challenge for Computer Vision. Traditional dense motion estimation algorithms have difficulties with non-rigid and large motions, both of which are frequently observed in smoke motion. We propose an algorithm for dense motion estimation of smoke. Our algorithm is robust, fast, and has better performance over different types of smoke compared to other dense motion estimation algorithms, including state of the art and neural network approaches. The key to our contribution is to use skeletal flow, without explicit point matching, to provide a sparse flow. This sparse flow is upgraded to a dense flow. In this paper we describe our algorithm in greater detail, and provide experimental evidence to support our claims.
\end{abstract}

\section{Introduction}

Accurate dense motion estimation is one of the fundamental problems in Computer Vision.  The problem is especially marked
for highly dynamic phenomena, such as smoke. Yet a solution for fluids is of value to a diverse set of areas of study. In Computer Graphics there are applications in both post-production \cite{gregson-SIGGRAPH2014,okabe-2012,ghanem-CVPR2008} and model acquisition \cite{li-2013}; In Atmospheric research, there are applications for storm identification and forecast~\cite{lakshmanan-2003stormforecast}, forecast and tracking the evolution of convective systems~\cite{vila-2008forecast} and rain cloud tracking~\cite{barbaresco-IP2001rain}. 

Our contribution is a method which gives simple, fast, robust, and accurate motion estimation of smoke. Our approach differs from most others because it do not rely on the constant brightness assumption, nor does it use feature matching of any kind. Smoke is partially transparent, it diffuses and aggregates as it moves. These characteristics breach the assumptions that typically underpin dense motion algorithms. This may explain why they under-perform in the case of smoke when compared to other object classes.

The key assumption in our motion estimation method is that the structure of the smoke density distribution does not change much from frame to frame. This allows us to estimate motion smoke in a global sense (described below) and thereby avoid the error-prone procedure of matching points in video frames based on local appearance, which is in constant flux. 

We now describe our approach briefly, considering a pair of video frames. Our algorithm estimates a dense flow using a sparse flow as a starting point. Our key assumption (global smoke density changes little from frame to frame) is responsible for the sparse estimation, which begins by assuming that pixel intensity and smoke density are directly proportional. In each frame, we construct a skeleton that characterises the smoke density. In this paper, we use density ridges as the basis for skeleton construction. We then estimate the motion of each skeletal point on skeleton one using all points on skeleton two. This gives a sparse flow that is upgraded to a dense flow by interpolation~\cite{garcia-CSDA2010,wang-EM2012} and refined using an existing energy minimisation approach~\cite{revaud2015epicflow}.

It is also worth noting that we do not use graph matching to map skeletons: graph matching is NP hard and it is difficult to account for any changes in structure that do occur. Local appearance information ({\em i.e.} similarity measures with features or patches as input) plays no role at all in our motion estimator: we have already observed appearance changes that may confuse standard motion estimation methods. Instead we independently map each skeletal point, $\mathbf{x}$, in skeleton one, $S_1$, in frame one to its expected location, $E[\mathbf{y}|\mathbf{x}] = \sum_{\mathbf{y} \in S_2} \mathbf{y}p(\mathbf{y}|\mathbf{x})$, when all points $\mathbf{y}$ in skeleton two, $S_2$, are considered.  

We describe our algorithm in detail in Section~\ref{sec:method}. Experiments against both classical and a range of state of the art alternatives (including latest CNN based methods) show our method outperforms all of them, see Section~\ref{sec:eval}. We wrap up the paper in Section~\ref{sec:conc} by pointing to both limitations and future applications of the work.

\section{Related Work}
\label{sec:relwork}

Optical flow pioneered by Horn and Schunck~\cite{hornschunck-taiu1981} is probably the best known and widely used general method for motion estimation. They assumed brightness constancy and local smoothness, leading to a regularised minimisation problem over the whole image. Although this method is generally applicable, it faces many challenges such as large motion displacement, motion discontinuities, {\em etc.}~\cite{butler-ECCV2012}. Advanced methods~\cite{brox2011large,mjblack-cviu1996,sun-CVPR2010} were proposed to solve the specific challenge in optical flow. With the deep learning trend, CNN based methods\cite{dosovitskiy2015flownet} are also proposed to deal with these challenges.

Unfortunately, none of these methods can properly deal with all of the many challenges that exist simultaneously in highly dynamic natural phenomena where the objects have no definitive features ({\em e.g.} smoke, water) or many similar features ({\em e.g.} leaves on trees). Brightness constancy is routinely violated in such cases, and if local features do exist they can move rapidly compared to their size. Furthermore, regularisors typically assume smoothness in the flow field; again an assumption that is often violated by natural phenomena. In Section~\ref{sec:eval}, we evaluate these methods with smoke cases.

Faced with these problems, the literature has seen a steady stream of papers that proposed solutions dedicated to the motion estimation of fluids. In this paper, we cannot mention all of them. Here we will discuss the most relevant and state of the art techniques. M\'{e}min and P\'{e}rez provide an early example~\cite{memin1999fluid}; they construct a dense flow by fitting parametric models to {\em critical points} (vortices of different classes) and interpolating between them.

In recent years, fluid motion estimation methods have been proposed that combine appearance based model with optical flow framework. These methods~\cite{auroux-EF2011,corpetti-PR2000,corpetti-PAMI2002,corpetti-EF2006} add constraints to prefer the fluid like motion in the energy minimization process. Corpetti~\textit{et al.}~\cite{corpetti-PR2000}, for example, apply a divergence-curl-type smoothness to replace the original smoothness term in optical flow framework. Auroux~\textit{et al.}~\cite{auroux-EF2011} propose to use a group of appearance feature based terms as candidates to replace the smoothness term. These methods, however, are all limited by the basic assumptions of optical flow.

In addition to appearance based fluid motion estimation methods, physical based fluid motion estimation methods are also proposed. Most of these methods~\cite{li-2013,anumolu-openfoam2012,li-cvpr2010recovering,doshi-IP2010}  are based on the Navier-Stoke(NS) equation to describe the fluid motion.
For example, Doshi~\textit{et al.} \cite{doshi-IP2010} proposed a process of optical flow based on the NS equation. However, this method only optimises for the smoothness of the flow field and ignores the accuracy of the estimation without an appropriate evaluation. Anumolu \textit{et al.} \cite{anumolu-openfoam2012} provided a framework for smoke simulation with strict physical regularities considering 'vorticity confinement' \cite{steinhoff-1994PhyofF}. Nevertheless, this framework only works for the interaction boundary between rigid object and smoke. Li \textit{et al.} \cite{li-cvpr2010recovering} proposed a brightness constancy constraint(BCC) based framework that combined the NS equations with the 3D potential flow. Although it could be extended to other physical models, this framework is not compatible with rotational flow and is limited by the physical model they used. 

Apart from the methods based on NS equations, many novel methods based on other physical properties have been proposed. For example, Sakaino \cite{sakaino-CVPR2008} proposed a motion estimation method based on the physical properties of waves. Those properties are components of a wave model which is generated by a combination of sinusoidal functions. Some recently proposed methods such as  ~\cite{ji-CVPR2013,xue-ECCV2014} are based on refractive properties, while \cite{ding-ICCV2011} is based on light path Snell's law. These methods, though, are limited to laboratory environments and require specific equipment configurations.

We provide an algorithm that makes general assumptions: No strong physical model is applied, instead, we assume that a skeletal structure can be used to characterise broad appearance and skeletal matching can be used to characterise broad motion. This general idea has been used by~\cite{barbaresco-IP2001rain} for rain cloud tracking with radar image; we differ in the way in which both construct and match skeletons. Noted that the proposed method is much simpler compare to~\cite{barbaresco-IP2001rain}. 


\section{Method}
\label{sec:method}

Our motion estimation method for smoke assumes that: (1) the density of smoke, $\rho$, at time $t$ is related to density at time $t+dt$ by
\begin{eqnarray}
\rho_{t+dt}(\mathbf{y}) = \int_{\mathbf{x} \in \Re^3} \phi(\mathbf{y},\mathbf{x})\rho_t(\mathbf{x})d\mathbf{x}
\end{eqnarray}
in which $\phi$ is a mass transfer function ($\rho_t(\mathbf{x})d\mathbf{x}$ being the mass) that includes systematic motion and diffusion; (2) that fluid motion is smooth if looked at globally, but possibly turbulent when looked at locally; and (3) the observed density of smoke and pixel intensity are in direct proportion to one another.
These assumptions are justified by observing smoke and other fluid in motion: we see a two dimensional projection of a three dimensional phenomena in which local motion details are superimposed on a global motion.

\begin{figure}[t!]
  \centering
  \includegraphics[width=0.99\linewidth]{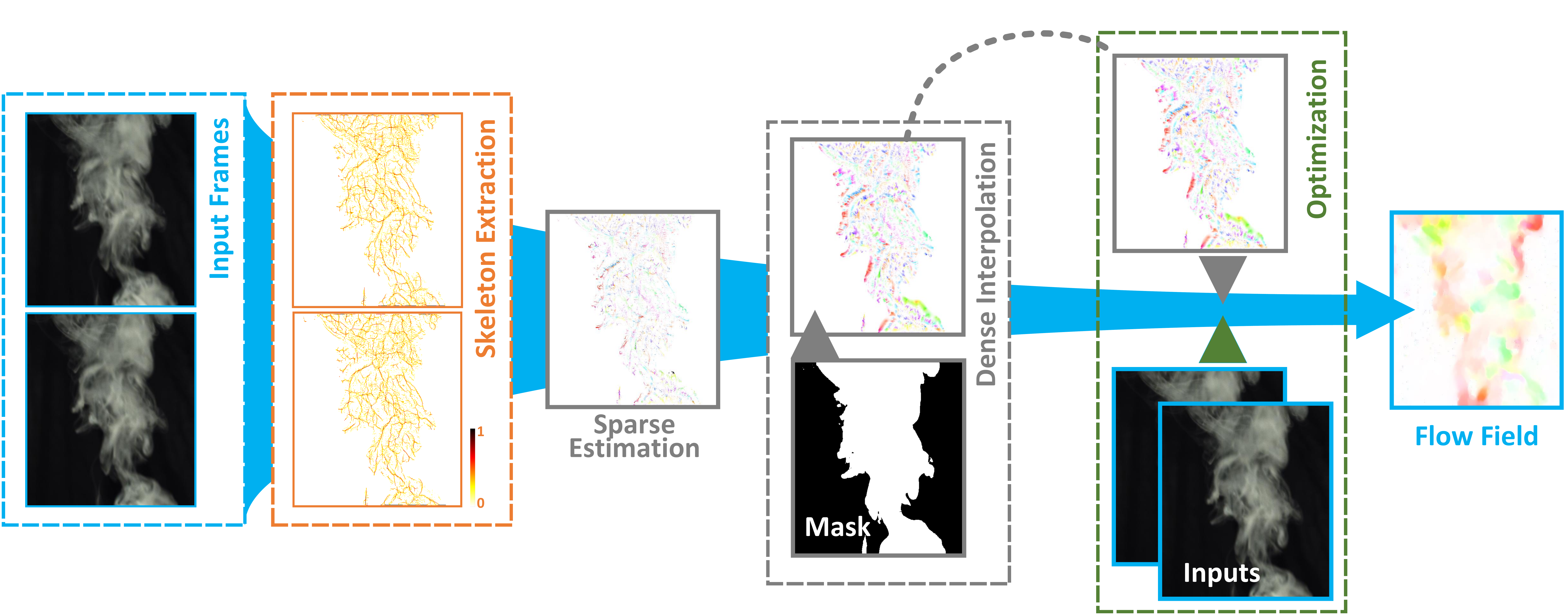}
\caption{System Overview. In our system pipeline, we first extract the smoke skeletons and estimate a sparse vector field in between (Section~\ref{sec:sparseEst}). This sparse information is then interpolated to a dense motion field (Section~\ref{sec:denseInterp}) which is further refined within a typical optical flow framework (Section~\ref{sec:oflowOpt}).}
\label{fig:system_structure}
\end{figure}

Ideally, we would like to solve for the transfer function $\phi(\mathbf{y},\mathbf{x})$, over all $\Re^3$, given a pair of input frames; but this is not possible. Instead we estimate the conditional density $p(\mathbf{y}|\mathbf{x})$, for a sparse set of  points  in $\Re^2$ that are chosen to characterise both the global and local motion of the flow, and then upgrade to a dense flow. In fact, our motion estimation algorithm has three main parts as shown in Figure~\ref{fig:system_structure}: (1) we produce a sparse estimation; (2) we interpolate the sparse estimation into a dense estimation; (3) we refine the dense estimation. Each part is now described in turn.

\subsection{Sparse Estimation}
\label{sec:sparseEst}

A novel sparse estimator is the key to our method and is the technical contribution of this paper. It is here that we approximate the transfer function with a conditional probability, for a sparse point set, in a manner that is sensitive to both global and local motion. Our sparse estimator has three substeps: (1) segment the smoke from the background; (2) fit a skeleton to each frame; (3) estimate sparse frame-to-frame motion using adjacent skeletons.

Algorithms exist that segment smoke and other natural phenomena from general backgrounds~\cite{Teney_etal_cvpr2015,chen-IP2013automatic_texture_segmentation,haindl-CAIP2013unsupervised_texture_segmentation,chan-cvpr2009variational_dynamic}. These methods provide various ways to detect areas with dynamic texture based on spatiotemporal filters~\cite{Teney_etal_cvpr2015}, optical flow motion field~\cite{chen-IP2013automatic_texture_segmentation}. But to focus on motion estimation alone, we use video acquired in laboratory conditions in which smoke appears to be light on a black background. This makes segmenting smoke simple: we threshold at some low value $\varepsilon $ to make a mask $M(\mathbf{x}) = f(\mathbf{x}) > \varepsilon $. For pixel values in $[0,255]$ we typically set $\varepsilon  = 1$.

The next step is skeleton fitting. The purpose of the skeleton is to characterise the density distribution of smoke particles so that global information can be used when obtaining a  sparse estimate for motion. The skeleton we use traces ridge lines of smoke density. It is a compound skeleton that is built from several individual skeletons, each characterising density in a different scale.

An individual skeleton comprises the ridge points of density (intensity) image that has been Gaussian blurred at scale $\sigma$. Let $g_\sigma = G_\sigma \odot f$ be the original image $f$ convolved with blur kernel $G_\sigma$. We locate ridges at pixels in $g_\sigma$ that are locally maximal in either the 'x' or 'y' direction, but we do not depend on this particular skeleton, we could also use~\textit{e.g.} a watershed skeleton~\cite{strahler-1957watershed}. The objective is to obtain a binary image $h_{\sigma}(\mathbf{x})$, which is a characteristic skeleton at scale $\sigma$. The skeleton is masked, using $M$, to ensure it lies only within the segmented smoke. The small scale skeletons capture local details, whereas the large scale skeletons capture global structure. 

\begin{figure}[t!]
  \centering
  \includegraphics[width=0.99\linewidth]{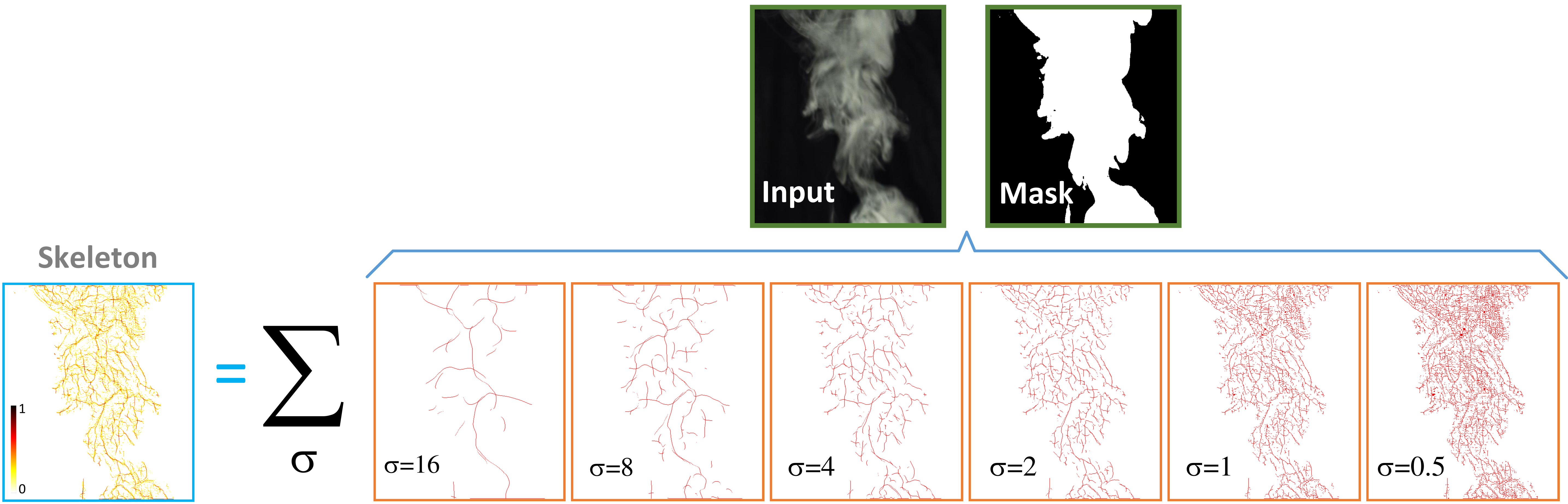}
\caption{Making a multi-valued smoke skeleton. Given an input smoke frame and a segmented mask, the skeleton can be aggregated by the sub-skeletons with different scale $\sigma$.}
\label{fig:smoke_skeleton}
\end{figure}

In principle it would be possible to use skeletons at different scales to obtain a motion estimate using a coarse-to-fine algorithm of some kind. We avoid that complication, instead preferring to build a compound skeleton by aggregation: 
\begin{equation}
h(\mathbf{x}) =  \frac{1}{N} \sum_{i=1}^{N} h_{\sigma_i}(\mathbf{x})
  \label{Skeleton}
\end{equation}
The result of this is a multi-valued skeleton that tends to emphasise stable structures within the smoke density: higher values of the skeleton indicate locations that are more stable over scale. Figure~\ref{fig:smoke_skeleton} illustrates the whole skeleton building process.

\begin{figure}[t!]
	\centering
	\includegraphics[width=0.9\linewidth]{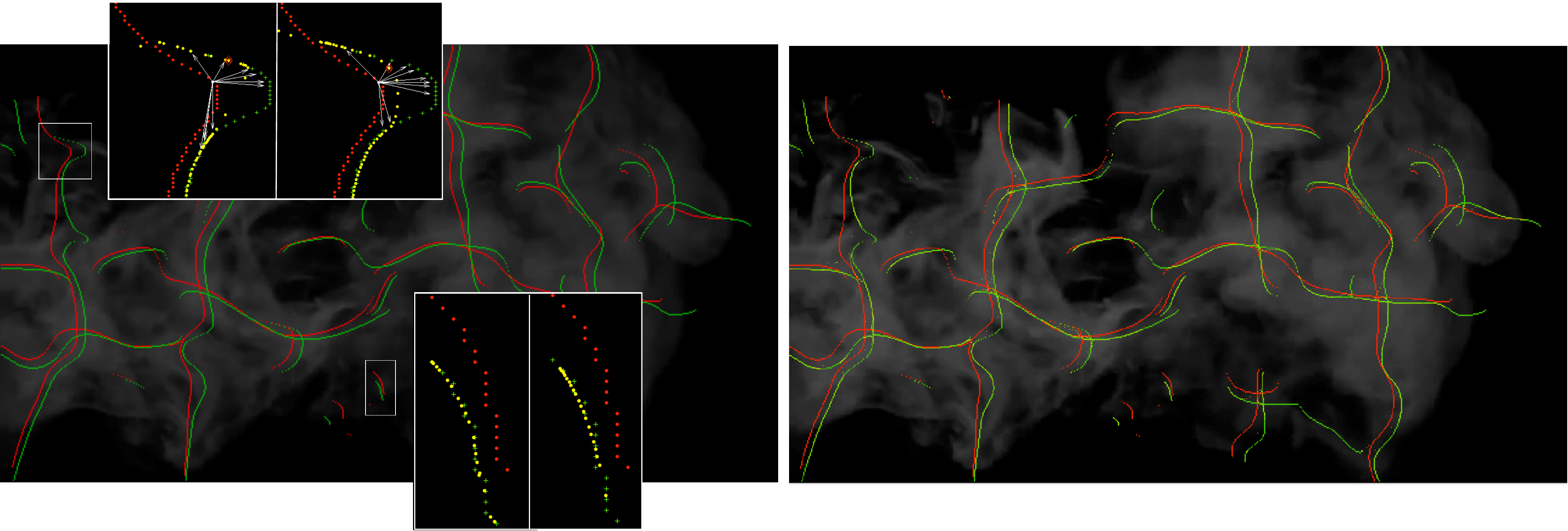}
	\caption{Left: frame one with starting skeleton (red) and attracting skeleton (green) in second frame. Detail at top  shows one point attracted to nearby points, expected motion is shown in yellow; right box shows the same, but for an isotropic Gaussian attractor. The bottom detail shows that an anisotropic Gaussian (left) leads to a better coverage of the attracting skeleton (green crosses) by the expected point destinations (yellow points) compared to an isotropic Gaussian (left). Right of figure shows the start skeleton (red), the attracting skeleton (green) and the sparse motion estimate (yellow).Please zoom in to see the details.}
\label{fig:pntatrct}
\end{figure}

With a multi-valued skeleton in hand, we can continue to estimate sparse motion between skeletons in adjacent frames.
Trying to match points 1-1 between skeletons is inappropriate: in principle because of the diffusion of smoke and in practice because the number of skeletal points is likely to change frame to frame. Iterative closest point~\cite{zhang-1994iterative} or similar technique to (for example) warp one skeleton into another requires the motion to be parametrisable: a strong assumption we wish to avoid. Using graph matching techniques is also not appropriate; graph matching is NP-hard, therefore pixels should not be nodes. The alternative is to link skeletal pixels into lines, but that is ad-hoc and complex.

Our solution is very simple. We specify the probability that a skeletal point $\mathbf{x}$ in frame at time $t$ has moved to skeletal point $\mathbf{y}$ in frame at time $t+dt$. This is a conditional probability, here denoted $p(\mathbf{y}|\mathbf{x})$. 
Then we compute the expected location at frame $t+dt$ of the point $\mathbf{x}$. For all $\mathbf{x} \in S_1$
\begin{eqnarray}
	E[\mathbf{y}|\mathbf{x}] = \sum_{\mathbf{y} \in S_2}
	 \mathbf{y}  p(\mathbf{y}|\mathbf{x}).
\end{eqnarray}
We note that this way of associating skeletal points $\mathbf{x}$ with destination points $\mathbf{y}$ imposes no strong constraints. There may be points $\mathbf{x}$ for which $\max p(\mathbf{y}|\mathbf{x}) < \epsilon$; these points can be filtered out as mapping to nothing. Conversely, there may be no point on the starting skeleton that maps to a given point on the destination skeleton, equally, many $\mathbf{x}$ could map to the same $\mathbf{y}$. Indeed, an expected point $E[\mathbf{y}|\mathbf{x}]$ is not constrained to lie on a skeleton. For these reasons we cannot properly describe our sparse estimator as being a matching algorithm, instead the destination skeleton is used as an attractor to construct a sparse flow estimation.

The conditional $p(\mathbf{y}|\mathbf{x})$ is defined using spatial distance and intensity of the skeletal pixels, similar to a bilateral filter~\cite{tomasi1998bilateral}
\begin{eqnarray}
p(\mathbf{y}|\mathbf{x}) \propto \mathcal{N}(\mathbf{x}|\mathbf{y},C_{y})
\mathcal{N}(h(\mathbf{x}) |h(\mathbf{y}),\sigma_v).
\end{eqnarray}
in which $h(.)$ is the value at a point in a multi-valued skeletal image. The standard deviation $\sigma_v$ is set to be related to the number of skeletal levels, typically $\sigma_v = 2/(N-1)$, where $N$ is the number of scales in skeleton searching. The covariance matrix $C_y$ is a non-isotropic Gaussian:
\begin{eqnarray}
 C_{y} &=&  ULU^T\\
 U &=& [\hat{\mathbf{n}},\ \hat{\mathbf{t}}]\\
 L &=& \sigma \mathrm{diag}( [1, \eta] )
\end{eqnarray}
in which $U$ is an orthonormal matrix that orients the Gaussian using the unit normal, $\hat{\mathbf{n}}$,  and unit tangent $\hat{\mathbf{t}}$ to the skeleton $S_2$ at $\mathbf{y}$.  The diagonal matrix $L$, squashes the distribution by a factor $\eta$ in the direction of the tangent to the skeleton, so that it narrows over the skeleton line. Typically $\eta = 1/10$. The reason for using an anisotropic, oriented Gaussian is to keep the distribution $\sum_{\mathbf{y} \in S_2} \mathcal{N}(\mathbf{z}|\mathbf{y},c_y)$ reasonably flat for all points $\mathbf{z} \in S_2$. 

If a non-isotropic Gaussian is used, the cumulative distribution will tend to peak in the middle of skeletal lines, with the result that expected values $E[\mathbf{y}|\mathbf{x}]$ will tend to crowd there also. This is seen in Figure~\ref{fig:pntatrct} which also illustrates that points in skeleton two attract points in skeleton one; skeletons are not explicitly matched.

It could be argued that we should make more use of the information in the multi-valued skeleton. In particular we could use the ``stability'' value, $v(\mathbf{x})$ in the multi-valued skeleton, similar to a coarse-to-fine strategy but ``stable to less-stable''. However, the use of a bilateral filter tends to make sure points in the starting skeleton are attracted to points of similar stability in the end skeleton; and given we are obtaining high quality results, this complication has not yet been pursued.

\subsection{Dense interpolation}
\label{sec:denseInterp}

Given a sparse flow estimation, the problem now is to upgrade this to a dense flow.  More exactly, we want to estimate a vector field $\mathbf{v}(\mathbf{x})$ for all points $\mathbf{x}$ in frame one using the partial estimate $S(\mathbf{x}) \cdot \mathbf{u}(\mathbf{x})$ in which  $S(\mathbf{x})$ is now a binary mask that locates points in the multi-valued skeleton, and $\cdot$ denotes the Hadamard product (also called Schur product or entrywise product); and $\mathbf{u}(\mathbf{x}) = E[\mathbf{y}|\mathbf{x}] - \mathbf{x}$ for all $\mathbf{x} \in S$.

We interpolate flow using a method due to Garcia {\em at al}~\cite{wang-EM2012}, which is designed for natural phenomenon. We consider each element of the vector field independently. Letting  $\mathbf{u} = [u_1,u_2]^T$ and $\mathbf{v} = [v_1,v_2]^T$ denote the fields, the interpolation process yields a new field $v_k$ by the following energy minimization problem:
\begin{equation}
\argmin_{v_k} \left \| M^{\frac{1}{2}} \cdot \left ( v_k - u_k \right )\right \|^{2} + \lambda \left \| \bigtriangledown ^{2} v_k \right \|^{2}
  \label{eqn:inter_energy}
\end{equation}
in which $M$ is a mask over the smoke pixels to indicate the smoke area for sparse to dense interpolation,  and $\lambda$ is a Lagrange multiplier that controls smoothness. Following Strang~\cite{strang-1999cosine_transform}, Garica~\cite{garcia-CSDA2010} shows that this least squares problem can be equivalently expressed 
using the discrete cosine transform (DCT) and its inverse (IDCT)
\begin{equation}
\textrm{IDCT}(\Gamma \cdot \textrm{DCT}(M \cdot (u_k - v_k)+ v_k))
  \label{inter_energy_rewrite}
\end{equation}
where $\Gamma$ is a 2D ``filtering tensor'' defined as:
\begin{equation}
\Gamma _{i_1,i_2} = \left ( 1+ \lambda \left ( \sum_{j=1}^{2}\left ( 2-cos\left(\frac{(i_j-1)\pi }{n_j}\right) \right ) \right )^{2} \right )^{-1}
  \label{inter_energy_filter_tensor}
\end{equation}
where $i_j$ locates the $i^{th}$ element along the $j^{th}$ dimension, and $n_j$ is the width of that dimension.
The algorithm proposed by Garica~\cite{garcia-CSDA2010} fixes the smoothing parameter, $\lambda$.

We recognise that this method of interpolation ignores any correlation between vector components. Even so our results are sufficiently good that we have opted to leave such matters aside for now.

\subsection{Optical Flow Energy and Optimization}
\label{sec:oflowOpt}

The dense flow estimated so far yields good results as shown in Section~\ref{sec:eval}. It shows improvement over all tested methods. Nevertheless, qualitative evaluation 
shows the field exhibits some noise. We now improve the motion estimation by smoothing it using variational optical flow energy, which requires we solve:

{
\centering
\begin{align}
E(\textbf{v}) &= \int_{\Omega} \underbrace{\phi ( \left \| f_1(\textbf{x}+\textbf{v}) - f_2(\textbf{x})\right \|^{2})}_{Brightness~Constancy}
+ \alpha \underbrace{\phi (\left \| \nabla f_{1}(\textbf{x}+\textbf{v}) - \nabla f_{2}(\textbf{x})\right \|^{2})}_{Gradient~Constancy}d\textbf{x}\nonumber\\
&+ \gamma \int_{\Omega} \underbrace{\phi(\left \| \nabla v_1 \right \|^{2} + \left \| \nabla v_2 \right \|^{2})}_{Smoothness~Constraint}d\textbf{x}
\end{align}
}

where $f_*$ denotes the input images and $\textbf{v}$ represents the smoothed flow field in between; $\nabla = (\partial_{xx},\partial_{yy})^{T}$ is a spatial gradient and $\phi(s)$ penalizes the flow gradient norm. The energy function was defined as a combination of a data term (brightness constancy and gradient constancy)~\cite{brox-ECCV2004} and a smoothness term~\cite{xu-PAMI2012}. 

In this case, given a full-size dense initial motion field, $\mathbf{v}(\mathbf{x})$, one level energy minimization is supposed to give better performance and precision comparing to the conventional coarse-to-fine scheme when it comes to the specific scenarios with boundary overlapping or thin objects~\cite{revaud2015epicflow}. Those difficulties often occur in smoke motion estimation. 

To optimize our proposed energy, we follow the same process in~\cite{revaud2015epicflow} by initializing the solution with our dense motion field from previous step and apply the fixed point iterations~\cite{brox-ECCV2004} without the coarse-to-fine scheme. The optimal flow field is obtained by solving the final linear systems using 30 iterations of the successive over relaxation method. All the parameters here are applied as same as~\cite{revaud2015epicflow}.

\section{Evaluation}
\label{sec:eval}

\begin{table}[!t]
\centering
\begin{tabularx}{1\textwidth}{c *{12}{Y}}
\toprule
IE & Ours & noEF & EF & HS & BA & NL & LDOF  & MDP & F\_C & F\_S\\
\midrule
 DLT & \textbf{0.093}    & 0.102            & 0.150    & 0.153 & 0.156 & 0.153    & 0.257 & 0.155 & 0.153      & \textit{0.148}      \\
 DMT & \textbf{0.116}    & \textbf{0.116}            & 0.301    & \textit{0.174} & 0.260 & 0.297    & 0.449 & 0.259 & 0.312      & 0.317      \\
LLT & \textbf{0.033}    & 0.040            & 0.052    & 0.054 & 0.054 & 0.053    & 0.053 & \textit{0.048} & 0.050      & 0.053      \\
 LMT & \textbf{0.072}    & 0.074            & 0.132    & \textit{0.102} & 0.103 & 0.106    & 0.113 & 0.111 & 0.106      & 0.115      \\
\bottomrule
\end{tabularx}
\vspace{0.5mm}
\caption{Quantitative Measure on Interpolation Error. We perform Interpolation Error (IE) metric on our smoke dataset (DLT, DMT, LLT and LMT) by comparing our method (Ours and noEF) to eight other baselines~\textit{i.e.} EpicFlow (EF), Horn\&Schunck (HS), Black\&Anandan (BA), Class+NL (NL), LDOF, MDP and FlowNet (F\_C and F\_S).}
\label{IE}
\end{table}

\begin{figure}[!t]
  \centering
  \includegraphics[width=0.9\linewidth]{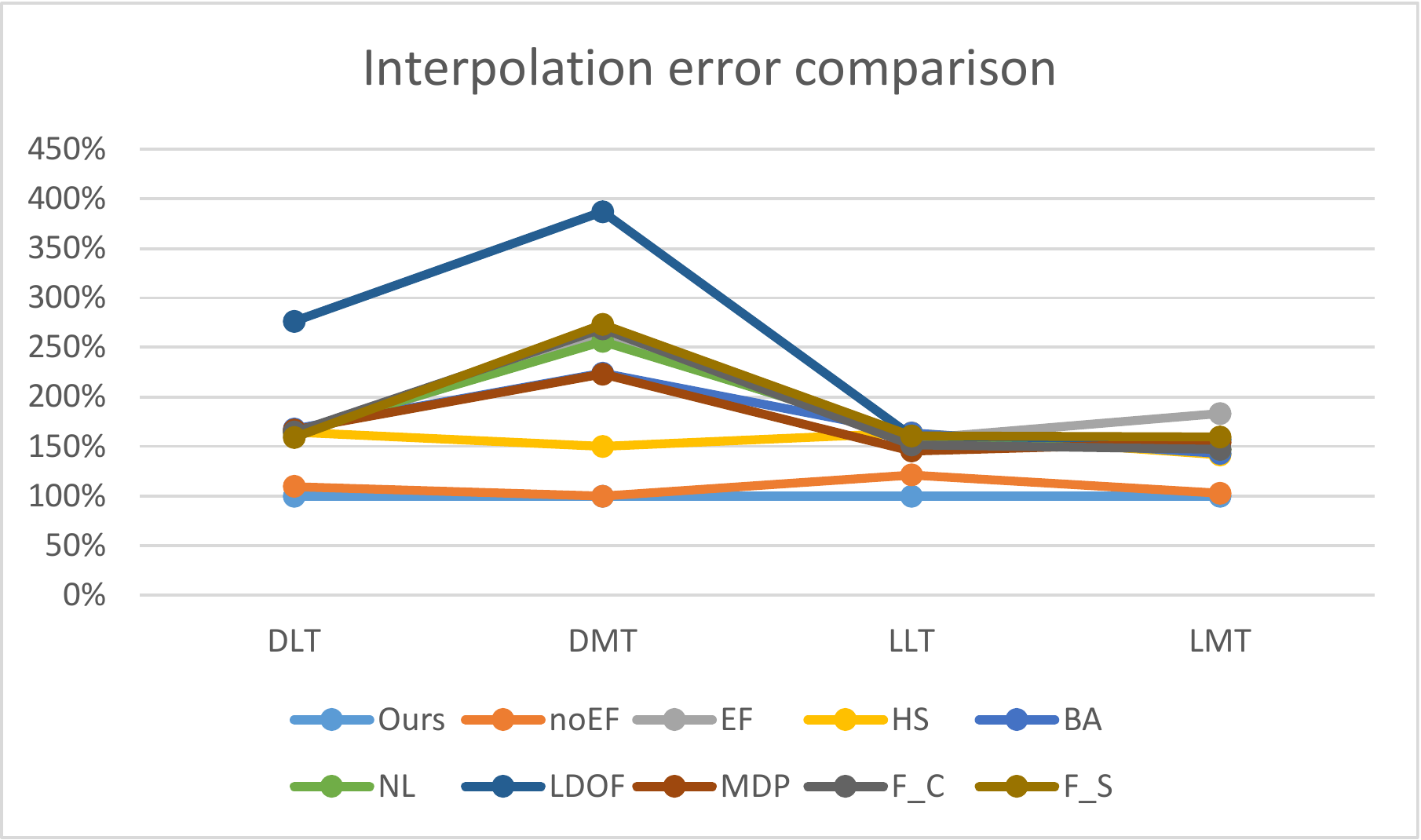}
\caption{Interpolation Error Comparison: Interpolation error for 4 test cases where all baselines are highlighted using different colours. The plotting is visually connected for the same algorithm. Noted that these results are all normalized by the best result shown in Table~\ref{IE}}
\label{IE_figure}
\end{figure}

\begin{figure}
  \centering
  \includegraphics[width=0.9\linewidth]{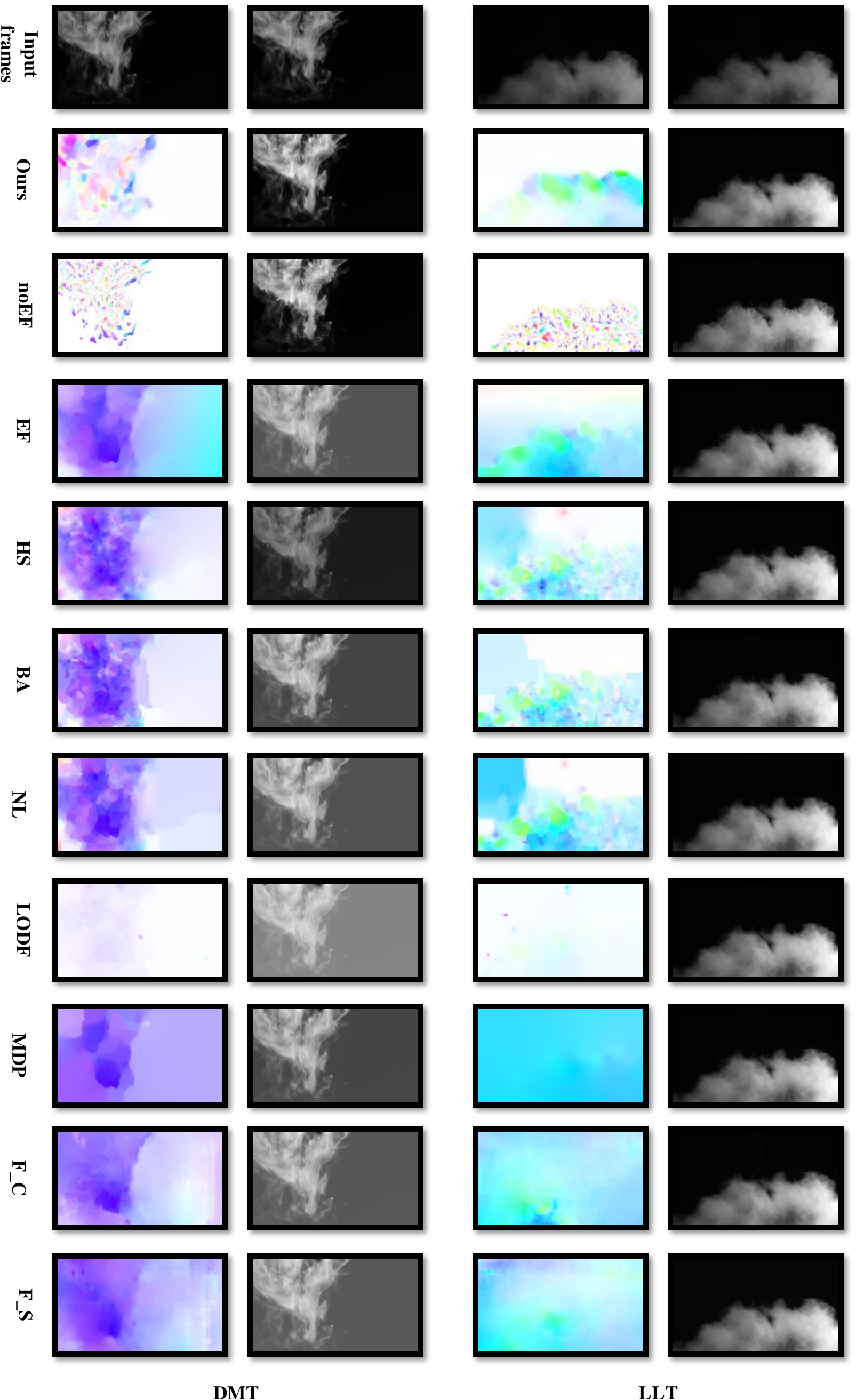}
\caption{Flow fields and warping results. The top row shows two sequences of input frames as DMT and LLT. Other figures are the results using all the methods in Table~\ref{IE}}
\label{fig:flow_field_warp}
\end{figure}

In this section, we evaluate our proposed method by comparing to eight baselines on four real-world sequences using the interpolation error (IE).

Given a ground truth benchmark~\cite{sbaker-ijcv2011,butler-ECCV2012}, the motion estimation algorithms are commonly evaluated using quantitative metrics~\textit{i.e.} endpoint error (EE) and angular error (AE). However, it is difficult to obtain ground truth motion field from the real-world smoke objects due to their transparency and high dynamic properties. To solve this issue, Baker~\textit{et al.}~\cite{sbaker-ijcv2011} provide another quantitative measure~\textit{i.e.} interpolation error(IE), for evaluating optical flow methods with a lack of ground truth. They applies the flow field result to warp the first input frame to estimate the second frame; then measure the interpolation error by comparing the warping result with the real second frame. 

Moreover, we consider the real-world sequence for all of our comparison because the synthetic smoke object is not able to fully reflect the dynamics of real-world motion behaviour. There is also no benchmark publicly available for providing real-world smoke data together with the ground truth motion. In this case, the interpolation error (IE) is supposed to be the only quantitative measure for such specific smoke scenario.

In the following measure, we compare two implementations (ours and noEF) of our method to both classic and the state-of-the-art motion estimation methods as follows. Horn and Schunck~\cite{hornschunck-taiu1981} (HS) is considered to be the first optical flow method in the field, while the Black and Anandan~\cite{mjblack-cviu1996} (BA) and Classic+NL~\cite{sun-CVPR2010} provide extra robustness to motion discontinuity by using motion regularization and median filtering respectively. Furthermore, Brox~\textit{et al.} (LDOF)~\cite{brox2011large} address the large motion displacement issue using the additional feature matching, while Xu~\textit{et al.} ~(MDP)~\cite{xu-PAMI2012} is currently one of the state-of-the-art methods with a leading performance on the Middleburry benchmark~\cite{sbaker-ijcv2011}. EpicFlow~\cite{revaud2015epicflow} and FlowNet~\cite{dosovitskiy2015flownet} further improve the performance by using the deep neural networks. Note that we do not consider the fluid motion estimators in context as the related source code is not publicly available to us.

\begin{figure}[t]
  \centering
  \includegraphics[width=0.9\linewidth]{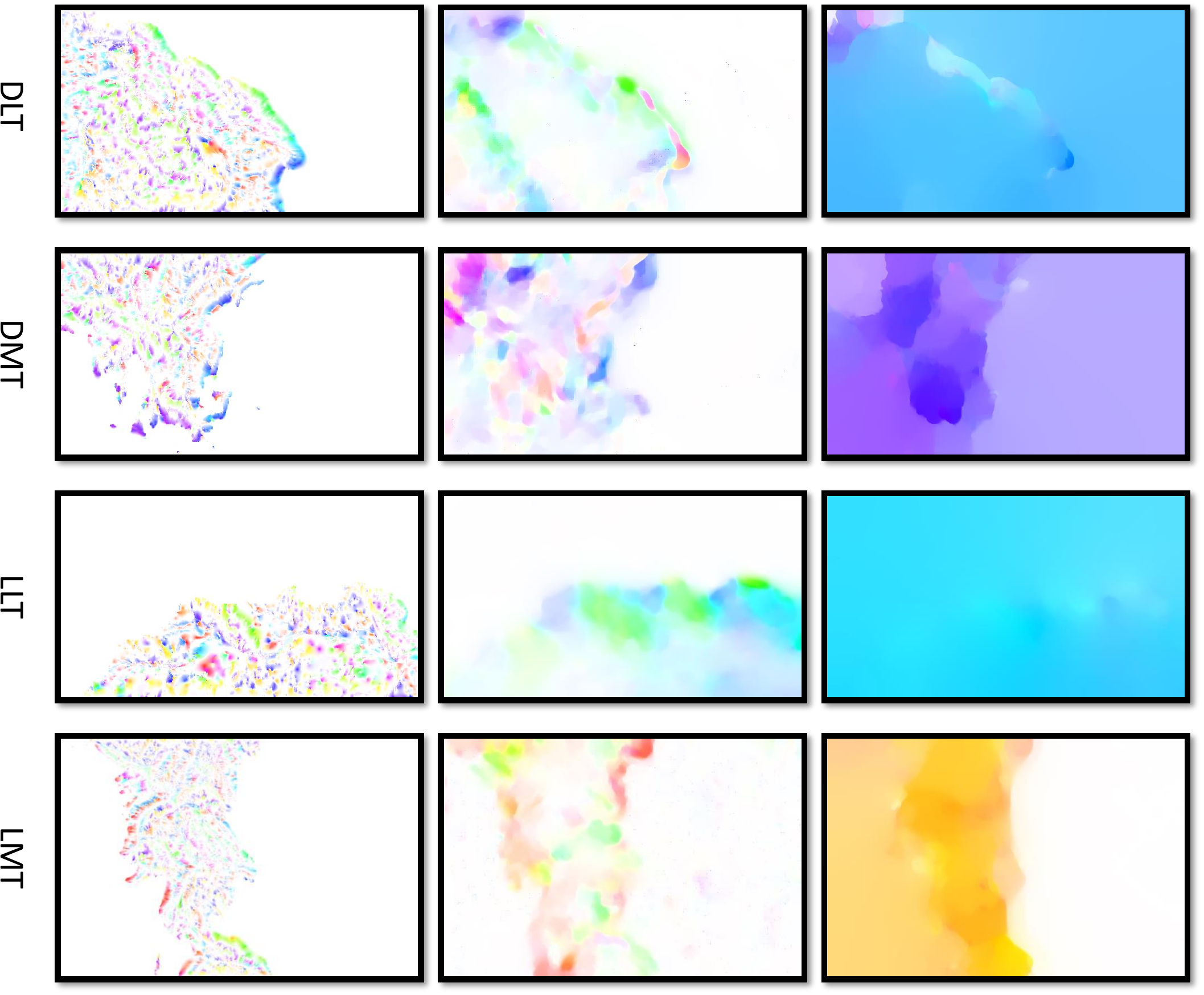}
\caption{Flow field comparison: Left: Proposed method without energy minimization. Middle: Proposed method with Energy minimization. Right: MDP method~\cite{xu-PAMI2012}}
\label{fig:flow field}
\end{figure}

\begin{figure}[t]
  \centering
  \includegraphics[width=0.9\linewidth]{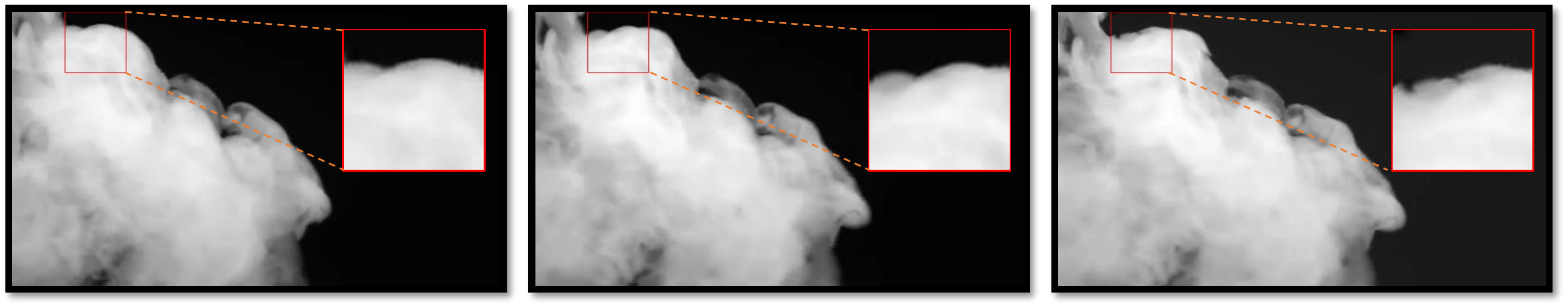}
\caption{Warping results comparison: Left: Ground truth frame Middle: Warping result using proposed method. Right: Warping result using MDP method~\cite{xu-PAMI2012}}
\label{fig:warping_results compare}
\end{figure}

To evaluate the performance of our algorithm, we picked four sequences that represent different real-world smoke \textit{i.e.} dense but low texture smoke (DLT), dense but more texture smoke (DMT), light low texture smoke (LLT) and light but more texture smoke (LMT). The quantitative comparison is presented in Table~\ref{IE} in which the first and second columns show our results with (Ours) and without (noEF) smoothing, respectively. In Figure~\ref{IE_figure}, we also visualize the IE results which are normalised by our best result. The proposed method outperforms all the tested approaches and significantly improves the motion precision in all different trials. On average, our result with smoothing is 149\% better than the second best state of art result and 108\% better than our result without smoothing. Figure~\ref{fig:flow_field_warp} shows the flow field and warping results for DMT and LLT using all the baselines. Our methods yield visually more realistic interpolation result as well as the preservation on the smoke boundary. 

Figure~\ref{fig:flow field} highlights a visual comparison of the flow fields produced by three different processes. Compared with MDP (a state-of-the-art), both of our proposed methods preserve the fine smoke motion details. By applying the flow field to warp the former input frame, the estimation results for the second frame are shown in Figure~\ref{fig:warping_results compare}. Compared with the second input frame, all the methods generate visually plausible structures of smoke since the structures between two input frames are similar. However, in the aspect of smoke detail capture, our algorithm outperforms MDP.

A key observation is that DMT is the hardest case for all methods based on Figure~\ref{IE_figure} and Table~\ref{IE}. We speculate this is because the DMT case contains lots of similar textures that confuse standard methods. Baselines such as EpicFlow(EF)~\cite{revaud2015epicflow} or BA~\cite{mjblack-cviu1996}, strongly rely on object edges. With complicated and deformable textures, the edge constraint is easily violated. Our skeletal method does not rely on texture or appearance. In both quantitative results and qualitative results, we are clearly better compared to all the other methods.


\section{Conclusion}
\label{sec:conc}

We have described a smoke motion estimation algorithm for various smoke types. It makes no assumptions about brightness, local appearance, or physics; it does assume global appearance is similar between frames. It assumes global appearance can be characterised by a skeleton, and 'skeleton matching' is used to construct a sparse flow that initiates a dense flow algorithm. Results on various kinds of smoke show we outperform a gamut of both classical and state of the art dense motion estimation methods, including CNN based methods.

The most significant limitation of the method is that it is two dimensional. However, the underlying mathematics of our sparse flow algorithm is not complicated and should extend to three dimensions. A second important limit is that so far we have used only video captured under laboratory conditions; but  work exists to segment smoke and other dynamic phenomena from real world scenes ({\em e.g.}
\cite{Teney_etal_cvpr2015}). Finally, we designed the motion estimation method for smoke and have not yet tested it on other highly dynamics objects such as fire.

Nonetheless, our motion estimation method is simpler and more robust than all algorithms we tested against, and it yields a model of smoke that is useful in Computer Graphics applications; an area we plan to explore.


\vspace{3mm}
\noindent {\bf Acknowledgement}. The authors are supported by the EPSRC project OAK EP/K02339X/1. We thank the reviewers for constructive comments, and also thank H. Gong for his helpful suggestion and comments.

\bibliographystyle{splncs}
\bibliography{ref}



\end{document}